\documentclass[]{spie}  

\usepackage{amsmath,amsfonts,amssymb}
\usepackage{graphicx}
\usepackage[colorlinks=true, allcolors=blue]{hyperref}
\usepackage{booktabs}
\usepackage{multirow}
\usepackage{enumitem}

\title{Do LLM-derived graph priors improve \\multi-agent coordination?}

\author[1]{Nikunj Gupta}
\author[2]{Rajgopal Kannan}
\author[1]{Viktor Prasanna}

\affil[1]{University of Southern California, Los Angeles, California, United States}
\affil[2]{DEVCOM ARL Army Research Office, Los Angeles, California, United States} 

\authorinfo{Further author information: (Send correspondence to Nikunj Gupta)\\Nikunj Gupta: E-mail: nikunj@usc.edu} 

\begin{document} 
\maketitle

\begin{abstract}
Multi-agent reinforcement learning (MARL) is crucial for AI systems 
that operate collaboratively in distributed and adversarial settings, 
particularly in multi-domain operations (MDO). A central challenge 
in cooperative MARL is determining how agents should coordinate: 
existing approaches must either hand-specify graph topology, rely on 
proximity-based heuristics, or learn structure entirely from 
environment interaction; all of which are brittle, semantically 
uninformed, or data-intensive. We investigate whether large language 
models (LLMs) can generate useful coordination graph priors for MARL 
by using minimal natural language descriptions of agent observations 
to infer latent coordination patterns. These priors are integrated 
into MARL algorithms via graph convolutional layers within a 
graph neural network (GNN)-based pipeline, and evaluated on four 
cooperative scenarios from the Multi-Agent Particle Environment 
(MPE) benchmark against baselines spanning the full spectrum of 
coordination modeling, from independent learners to state-of-the-art 
graph-based methods. We further ablate across five compact open-source 
LLMs to assess the sensitivity of prior quality to model choice. 
Our results provide the first quantitative evidence that LLM-derived 
graph priors can enhance coordination and adaptability in dynamic 
multi-agent environments, and demonstrate that models as small as 
1.5B parameters are sufficient for effective prior generation. 

\end{abstract}


\section{Introduction} 
\label{sec:introduction} 
Multi-agent reinforcement learning (MARL) has emerged as a foundational 
paradigm for training collections of autonomous agents to act, 
communicate, and coordinate within shared, dynamic environments. Its 
applications span a wide and growing spectrum: autonomous vehicle 
fleets, UAV swarms~\cite{tahir2019swarms, alqudsi2025uav}, warehouse 
robotics~\cite{zhen2025optimizing, keith2024review}, distributed sensor 
networks~\cite{mason2024multi}, human-robot teams~\cite{senaratne2025framework}, 
and safety-critical military contexts such as multi-domain operations 
(MDO)\cite{townsend2018accelerating,skates2021multi,wojtowicz2018multi}. In all 
of these settings, the ability of agents to coordinate 
(i.e., to form coherent joint strategies despite acting on local, 
partial observations) is not merely desirable but essential. Yet 
coordination remains one of the central open challenges in MARL. 
Agents must implicitly or explicitly reason about the intentions and 
observations of their teammates, often under tight computational and 
communication constraints~\cite{bernstein2002complexity, 
oliehoek2016concise}. Classical approaches such as independent 
Q-learning~\cite{tan1993multi} ignore inter-agent dependencies 
entirely, while fully centralized methods suffer from exponential 
growth in the joint action space~\cite{zhang2021multi, 
oroojlooy2023review}. The centralized training with decentralized 
execution (CTDE) paradigm~\cite{kraemer2016multi, amato2024introduction} 
has offered a principled middle ground, yielding influential value 
decomposition algorithms such as VDN~\cite{sunehag2017value}, 
QMIX~\cite{rashid2020monotonic}, QTRAN~\cite{son2019qtran}, and 
QPLEX~\cite{wang2020qplex}. While effective, these methods treat 
the agent team as an undifferentiated collective, failing to 
explicitly model which agents should attend to which 
others. Graph-based MARL addresses this by encoding agent relationships 
as edges in a coordination graph, using GNNs~\cite{veličković2018graph} 
to propagate information along these edges and ground value 
factorization in explicit relational 
structure~\cite{bohmer2020deep, li2020deepimplicit, wang2022contextaware, 
kang2022non, yang2022self, duan2024group, gupta2025deep}. However, 
existing approaches must either hand-specify the graph topology, rely 
on proximity-based heuristics, or learn structure entirely from 
environment interaction, all of which are brittle, semantically 
uninformed, or data-intensive. This raises a natural question: can a 
richer, semantic source of prior knowledge be leveraged to initialize 
coordination graphs before learning begins?

\noindent Large language models (LLMs) have simultaneously undergone a remarkable 
rise as general-purpose reasoning engines. Both large proprietary models 
and increasingly capable open-source alternatives have demonstrated 
strong performance in chain-of-thought reasoning~\cite{wei2022chain}, 
common-sense inference, strategic planning~\cite{yao2023react}, and 
zero-shot policy generation for sequential decision-making 
tasks~\cite{huang2022language, wang2023voyager}. These capabilities 
are particularly relevant to multi-agent coordination for several 
reasons. First, LLMs encode vast prior knowledge about teamwork, 
roles, and strategy, distilled from human text spanning game theory, 
military doctrine, organizational behavior~\cite{mooney1947principles}, 
and collaborative problem-solving. Second, their strong natural 
language understanding enables them to interpret semantically rich 
descriptions of agent observations, such as unit types, relative 
positions, health states, and task objectives, and infer meaningful 
relational structure from them, a task that purely data-driven methods 
struggle with under limited experience~\cite{sun2024llmmarl}. Third, 
LLMs can perform this inference in a zero-shot fashion, 
requiring no environment-specific training or labeled coordination 
data~\cite{huang2022language}. Together, these properties suggest 
that LLMs may be uniquely well-positioned to serve not as direct 
policy executors~\cite{slumbers2024leveraging}, but as structural prior generators or sources of 
high-level coordination knowledge that can be injected into a 
downstream learning system to guide and accelerate policy optimization.  

\begin{figure*}[t]
    \centering
    \includegraphics[width=\textwidth]{figures/introduction_figure.pdf}
    \caption{MARL applications and the coordination challenge. Agents 
    across domains must coordinate under partial observability. We 
    propose LLM-derived graph priors as a semantic, data-efficient 
    alternative to heuristic or learned coordination graphs.}
    \label{fig:intro}
\end{figure*}

\noindent In this paper, we investigate precisely this hypothesis. Specifically, 
we study whether LLMs can generate useful coordination graph 
priors for GNN-based MARL, and whether doing so leads to measurable 
improvements in fully cooperative task performance. Given minimal 
natural language descriptions of each agent's local observation, we 
prompt an LLM to infer which agents are likely to benefit from 
coordinating with one another. The LLM's output is parsed into a 
weighted adjacency matrix, which serves as a graph prior over the 
agent coordination topology. This prior is consumed by a graph neural 
network (GNN)~\cite{veličković2018graph} that aggregates agent 
observations and produces enriched feature representations, which in 
turn feed into a standard MARL policy optimization pipeline under the 
CTDE paradigm. We evaluate this framework on four cooperative 
scenarios from the Multi-Agent Particle Environment 
(MPE)~\cite{lowe2017multi, mordatch2017emergence}, comparing against 
baselines spanning the full spectrum of coordination modeling and 
ablating across five compact open-source LLMs to assess prior quality 
sensitivity to model choice.

\noindent The contributions of this paper are as follows:

\begin{itemize}[nosep]
    \item \textbf{LLM-driven coordination graph prior generation.} 
    We propose a method for eliciting structured coordination graph 
    priors from LLMs via natural language prompting of agent 
    observations, requiring no task-specific fine-tuning and no 
    ground-truth graph supervision, making it broadly applicable 
    across cooperative MARL settings.

    \item \textbf{Empirical evaluation on MPE.} We demonstrate that 
    LLM-derived graph priors improve policy performance across four 
    cooperative MPE environments, providing systematic comparisons 
    against baselines spanning four categories of coordination 
    modeling, from independent learning to state-of-the-art 
    soft-edge graph methods~\cite{bohmer2020deep, li2020deepimplicit, 
    gupta2025deep, gupta2026action}.

    \item \textbf{LLM sensitivity and deployability analysis.} We 
    ablate across five compact open-source models (0.5B--3B 
    parameters) spanning three model families, demonstrating that 
    models as small as 1.5B parameters are sufficient for effective 
    coordination graph prior generation, and providing practical 
    guidance for resource-constrained multi-agent deployment.
\end{itemize}

\noindent The remainder of this paper is organized as follows. 
Section~\ref{sec:related} reviews related work in MARL, GNN-based 
coordination, and LLMs for decision-making. 
Section~\ref{sec:Methodology} describes our proposed framework. 
Section~\ref{sec:Experiments} presents our experimental setup and 
results on MPE. Section~\ref{sec:Discussion} discusses broader 
implications, limitations, and future directions. 
Section~\ref{sec:Conclusion} concludes. 

\section{Background and Related Works} 
\label{sec:related} 

\subsection{Multi-Agent Reinforcement Learning} 

Cooperative multi-agent reinforcement learning (MARL) studies how a team of agents can learn decentralized decision-making policies that maximize a shared objective through interaction with a common environment. This setting is commonly formalized as a Decentralized Partially Observable Markov Decision Process (Dec-POMDP)~\cite{bernstein2002complexity, oliehoek2016concise}, defined by the tuple
\begin{equation}
\mathcal{M} = \langle \mathcal{I}, \mathcal{S}, \{\mathcal{A}_i\}_{i\in\mathcal{I}}, \mathcal{T}, \mathcal{R}, \{\mathcal{O}_i\}_{i\in\mathcal{I}}, \mathcal{O}, \gamma, \rho_0 \rangle,
\end{equation}
where $\mathcal{I}=\{1,\dots,n\}$ is the set of agents, $\mathcal{S}$ is the global state space, $\mathcal{A}_i$ is the local action space of agent $i$ (with joint action space $\mathcal{A}=\prod_{i\in\mathcal{I}}\mathcal{A}_i$), $\mathcal{T}(s' \mid s, \mathbf{a})$ is the transition function, $\mathcal{R}(s,\mathbf{a})$ is a shared reward function, and $\gamma \in (0,1)$ is the discount factor. Each agent receives a local observation $o_i \in \mathcal{O}_i$, with joint observation space $\mathcal{O}=\prod_{i\in\mathcal{I}}\mathcal{O}_i$, and the environment is initialized from $\rho_0$. At execution time, agent $i$ selects actions using only its local action--observation history, which makes coordination under partial observability the central challenge in cooperative MARL. 

\noindent One of the dominant frameworks for learning in Dec-POMDPs is centralized 
training with decentralized execution (CTDE) \cite{kraemer2016multi, 
amato2024introduction}. Under CTDE, global information (e.g., state or 
joint observations) may be used during training to stabilize learning, 
while each agent deploys a decentralized policy that conditions only on 
local information. Value decomposition is a prominent CTDE family: 
methods such as VDN~\cite{sunehag2017value}, 
QMIX~\cite{rashid2020monotonic}, Weighted 
QMIX~\cite{rashid2020weighted}, QTRAN~\cite{son2019qtran}, and 
QPLEX~\cite{wang2020qplex} factorize a centralized $Q_{\mathrm{tot}}$ 
into per-agent utilities under structural constraints that permit 
tractable decentralized action selection. Complementary actor-critic 
extensions under CTDE, including MADDPG~\cite{lowe2017multi} and 
MAPPO~\cite{yu2022mappo}, extend this paradigm to policy gradient 
methods, enabling stable training with continuous action spaces and 
richer policy representations. More recent work has explored sequence 
modeling perspectives on MARL~\cite{wen2022multi}, conditional policy 
factorization~\cite{wang2023more}, correlated policy 
structures~\cite{li2025agentmixer} and uncertainty-aware 
action modeling~\cite{gupta2023cammarl} to further improve the expressiveness 
of joint value functions within the CTDE framework. Beyond algorithmic 
design, orthogonal lines of work have addressed the computational 
scalability of MARL training on modern multi-core and multi-GPU 
hardware~\cite{wiggins2025arc, wiggins2025accelerating}, reflecting 
the growing practical demands of deploying cooperative MARL at scale. 


\noindent While these approaches achieve strong empirical performance, they often treat the agent team as a homogeneous collective and do not explicitly encode which agents should coordinate most strongly. In many tasks, coordination is structured and context-dependent: only subsets of agents may need tight coupling, and these dependencies may change with roles, spatial configuration, or information asymmetry. Graph-based MARL addresses this by representing agents as nodes and learning relational interactions through graph neural networks (GNNs), but most existing approaches rely on heuristic graph constructions (e.g., fully connected or proximity graphs) or learn edges purely from low-level state features. In this work, we investigate whether Large Language Models (LLMs) can provide a complementary source of structural inductive bias by generating coordination graph priors from high-level semantic descriptions of agent observations.

\subsection{Graph-Based and GNN-Driven MARL}

Graph neural networks provide a natural inductive bias for multi-agent systems, 
as agent teams are inherently relational: the value of an agent's action often 
depends critically on the actions of a subset of its teammates rather than the 
full team. Coordination graphs (CGs) formalize this intuition by representing 
agents as nodes and coordination dependencies as edges, factorizing the joint 
value function over local agent 
interactions~\cite{bohmer2020deep, li2020deepimplicit}. Deep Coordination 
Graphs (DCG)~\cite{bohmer2020deep} learn graph structure and value functions 
end-to-end via message passing, while Deep Implicit Coordination Graphs 
(DICG)~\cite{li2020deepimplicit} infer soft graph structure implicitly from 
agent observations using attention mechanisms~\cite{veličković2018graph}. 

\noindent Subsequent work has explored richer graph topologies and more expressive 
factorizations. Context-Aware Sparse Deep Coordination 
Graphs~\cite{wang2022contextaware} learn sparse, context-dependent edges to 
reduce communication overhead. Non-Linear Coordination 
Graphs~\cite{kang2022non} relax the linearity assumptions of earlier 
factorizations to improve representational capacity. Self-Organized 
Polynomial-Time Coordination Graphs~\cite{yang2022self} address the 
computational cost of exact inference over dense graphs. Group-Aware 
Coordination Graphs~\cite{duan2024group} introduce hierarchical agent grouping 
to capture multi-scale coordination structure. Graph-based communication strategies~\cite{ruan2022gcs, liu2023deep, gupta2025hammer} further extend the graph paradigm to govern structured 
message passing and multi-level coordination during decentralized execution 
itself, enabling agents to selectively share information based on learned 
relational importance. More recently, Deep Meta Coordination Graphs~\cite{gupta2025deep}, 
TIGER-MARL~\cite{gupta2025tiger}, and Action-Graph 
Policies~\cite{gupta2026action} have introduced temporally-aware and 
action-dependency-aware dynamic graph structures that adapt to evolving 
agent configurations and task contexts. 
A broader survey of graph-based reinforcement learning for networked 
coordination is provided in~\cite{liu2025survey}.

\noindent A key limitation shared by virtually all of these methods is that the 
coordination graph must be hand-specified, constructed via simple heuristics 
such as agent proximity, or learned entirely from environment interaction. 
None of these approaches leverage external semantic knowledge about agent 
roles and relationships prior to learning. Our work directly addresses this gap. 

\subsection{Large Language Models for Decision-Making}

Large language models (LLMs) have emerged as powerful general-purpose 
reasoning engines, demonstrating strong capabilities in chain-of-thought 
reasoning, common-sense inference, and zero-shot generalization across 
diverse domains. In single-agent sequential decision-making, LLMs have 
been applied as zero-shot planners that decompose high-level task 
descriptions into sequences of executable actions without any 
task-specific training~\cite{huang2022language}, as reasoning-and-acting 
agents that interleave verbal reasoning traces with environment 
interactions~\cite{yao2023react}, as open-ended skill learners in 
complex world environments~\cite{wang2023voyager}, and as reward shapers 
that provide dense feedback signals to guide policy 
optimization~\cite{eureka2023ma}. These works collectively establish 
that LLMs can serve as flexible sources of task knowledge for 
reinforcement learning agents and not merely as language processors, but 
as reasoning modules that translate semantic task understanding into 
actionable structure. 

\noindent The extension of LLM-based reasoning to {multi-agent} settings 
is a more recent and still nascent development~\cite{sun2024llmmarl}. 
A growing body of work has explored using LLMs to generate agent 
communication protocols, assign roles, produce high-level strategic 
plans, and coordinate behavior in cooperative multi-agent 
systems~\cite{slumbers2024leveraging}. Frameworks such as FAMA align 
LLMs with cooperative task objectives via centralized MARL update rules 
and exploit natural language for inter-agent 
communication~\cite{slumbers2024leveraging}. Other approaches treat 
LLMs as orchestrators that dynamically direct and sequence the actions 
of subordinate agents in response to evolving task 
states~\cite{neurips2025orchestration}. The coordinated routing of 
specialized LLMs via reinforcement learning has also been 
explored~\cite{gupta2025hierrouter}, highlighting the broader potential 
of combining LLM-based reasoning with learned decision-making frameworks. 
Despite this progress, the predominant paradigm in LLM-based MARL 
treats LLMs as {policy-level supervisors} or 
{communication facilitators}, i.e., agents that produce natural 
language actions or mediate inter-agent messaging. What remains largely 
unexplored is the use of LLMs as sources of {structural} prior 
knowledge about agent coordination topology: not to act or communicate, 
but to inform {which} agents should be relationally coupled 
before learning begins. Our work directly addresses this gap.

\noindent A practically important consideration for integrating LLMs into MARL 
pipelines is the computational cost of inference. Large proprietary 
frontier models introduce prohibitive latency and API costs that are 
incompatible with real-time multi-agent decision loops and distributed 
deployment scenarios. This motivates the use of compact, open-source 
alternatives that can be deployed locally without network dependencies. 
Recent model families including Qwen2.5~\cite{qwen2025}, 
LLaMA~\cite{grattafiori2024llama3}, and Gemma~\cite{gemma2024} have 
demonstrated competitive reasoning and instruction-following performance 
at parameter counts of 0.5B--3B, making them attractive candidates for 
resource-constrained multi-agent applications such as edge-deployed 
robotic systems and MDO~\cite{townsend2018accelerating,skates2021multi,wojtowicz2018multi}. 
In this work, we exclusively evaluate compact open-source models and 
explicitly analyze the tradeoff between model capability and 
computational overhead in the context of coordination graph prior 
generation. 




\subsection{Problem Setting}

We consider a fully cooperative Dec-POMDP with $n$ agents. At each timestep 
$t$, agent $i$ receives a local observation $o_i^t$ which can be described 
in natural language as a short textual summary $s_i^t$. We define a 
{coordination graph prior} as a weighted adjacency matrix 
$\mathbf{A} \in \mathbb{R}^{n \times n}$, where $\mathbf{A}_{ij}$ reflects 
the degree to which agent $i$ should attend to agent $j$ during coordination. 
Our goal is to use an LLM to generate $\mathbf{A}$ from the set of natural 
language observation summaries $\{s_1, \ldots, s_n\}$, and to integrate this 
prior into a GNN-based MARL pipeline to improve cooperative task performance. 
This formulation is agnostic to the underlying MARL algorithm and requires 
no task-specific fine-tuning of the LLM. 

\section{Methodology} 
\label{sec:Methodology} 

\subsection{Overview}

\begin{figure*}[t]
    \centering
    \includegraphics[width=\linewidth]{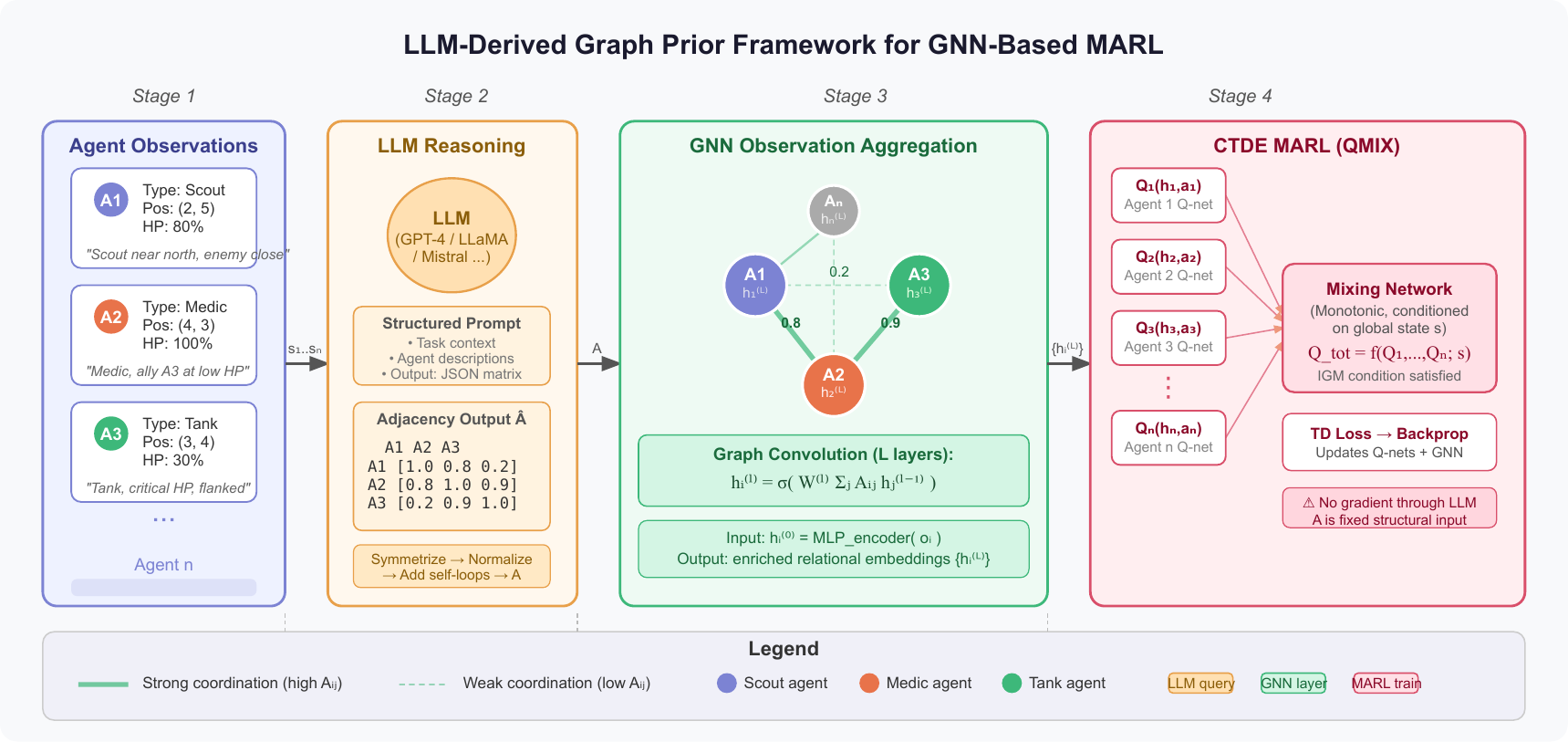}
    \caption{Overview of the proposed framework. Natural language descriptions 
    of agent observations are passed to an LLM, which produces a coordination 
    graph prior as a weighted adjacency matrix. This matrix is consumed by a 
    GNN that aggregates agent observation embeddings, producing enriched 
    relational feature representations that feed into a standard CTDE-based 
    MARL policy optimization pipeline.}
    \label{fig:architecture}
\end{figure*}

Our proposed framework introduces LLM-derived coordination graph priors into 
a GNN-based MARL pipeline. The full pipeline, illustrated in 
Figure~\ref{fig:architecture}, consists of four stages: (1) natural language 
observation description, (2) LLM-based graph prior generation, (3) GNN-based 
observation aggregation, and (4) MARL policy optimization under the CTDE 
paradigm. Each stage is described in detail below.

\subsection{Natural Language Observation Descriptions}

At each environment reset (or at a fixed re-prompting interval during 
execution) each agent $i$ produces a short natural language description 
$s_i$ of its local observation $o_i$. These descriptions capture task-relevant 
semantic information such as nearby entity types, relative positions, agent 
roles, and current objective states. Formally, we define a deterministic 
description function $\phi: \mathcal{O}_i \rightarrow \mathcal{S}$, where 
$\mathcal{S}$ is the space of natural language strings, such that:
\begin{equation}
    s_i = \phi(o_i).
\end{equation}
The description function $\phi$ is hand-authored once per environment and 
requires no learning. It maps structured observation vectors to concise 
natural language summaries, for example: \textit{``Agent 2 is a scout unit 
located near the northern boundary, observing two enemy units at close range 
and one allied support unit at medium range.''} Importantly, $\phi$ does not 
require access to the global state and operates entirely on each agent's 
local observation, preserving the decentralized nature of the system.

\subsection{LLM-Based Coordination Graph Prior Generation}

Given the set of natural language observation descriptions 
$\mathcal{S} = \{s_1, s_2, \ldots, s_n\}$, we construct a structured prompt 
$\mathcal{P}(\mathcal{S})$ and query an LLM to infer a coordination graph 
prior. The prompt instructs the LLM to reason about which agents are likely 
to benefit from coordinating with one another given their current observations, 
and to return a weighted adjacency matrix $\mathbf{A} \in [0,1]^{n \times n}$ 
encoding these beliefs, where $\mathbf{A}_{ij}$ reflects the coordination 
affinity between agent $i$ and agent $j$.

\subsubsection{Prompt Design}

The prompt $\mathcal{P}(\mathcal{S})$ is constructed to elicit structured coordination reasoning from the LLM while maintaining consistency across models. First, we provide a concise system-level description of the cooperative task and the functional role of the agents. This contextual grounding enables the LLM to reason about potential coordination dependencies in a task-aware manner. Next, we present the complete set of natural language observation descriptions $\{s_1, \ldots, s_n\}$, each explicitly indexed by its corresponding agent identifier to preserve relational clarity. Finally, we include a strict output instruction directing the LLM to return only a valid $n \times n$ JSON matrix with real-valued entries in the range $[0, 1]$, where larger values indicate stronger coordination affinity. The instruction further specifies that the matrix must be symmetric, i.e., $\mathbf{A}_{ij} = \mathbf{A}_{ji}$, ensuring compatibility with undirected graph representations. Unless otherwise noted, we do not use few-shot demonstrations or chain-of-thought prompting, allowing us to evaluate the zero-shot structural inference capabilities of each LLM under a controlled and uniform prompting regime.

\subsubsection{Graph Prior Extraction}

The LLM response is parsed to extract a raw adjacency matrix $\hat{\mathbf{A}} \in \mathbb{R}^{n \times n}$ representing inferred coordination affinities. To ensure suitability for downstream graph neural network (GNN) integration, we apply a sequence of post-processing operations. First, we enforce symmetry by computing $\mathbf{A} = \frac{1}{2}(\hat{\mathbf{A}} + \hat{\mathbf{A}}^\top)$, thereby producing an undirected interaction structure. Second, we perform row-wise normalization such that $\mathbf{A}_{i,:} \leftarrow \mathbf{A}_{i,:} / \sum_j \mathbf{A}_{ij}$, which stabilizes message aggregation during GNN propagation. Finally, we incorporate self-loops by updating $\mathbf{A} \leftarrow \mathbf{A} + \mathbf{I}$, ensuring that each agent retains access to its own feature representation during graph-based information integration. The resulting matrix $\mathbf{A}$ constitutes the final LLM-derived coordination graph prior used within the MARL framework.

\subsection{GNN-Based Observation Aggregation}

The coordination graph prior $\mathbf{A}$ is consumed by a graph neural 
network that aggregates agent observation embeddings into enriched relational 
feature representations. Each agent $i$ first produces a local embedding 
$\mathbf{h}_i^{(0)} \in \mathbb{R}^d$ from its raw observation $o_i$ via a 
shared MLP encoder $f_\theta$:
\begin{equation}
    \mathbf{h}_i^{(0)} = f_\theta(o_i).
\end{equation}
We then apply $L$ layers of graph convolution over the agent graph defined 
by $\mathbf{A}$. At each layer $\ell$, the embedding of agent $i$ is updated 
by aggregating the embeddings of its neighbors weighted by the coordination 
prior:
\begin{equation}
    \mathbf{h}_i^{(\ell)} = \sigma\left( \mathbf{W}^{(\ell)} 
    \sum_{j=1}^{n} \mathbf{A}_{ij} \, \mathbf{h}_j^{(\ell-1)} \right),
\end{equation}
where $\mathbf{W}^{(\ell)} \in \mathbb{R}^{d \times d}$ are learnable weight 
matrices and $\sigma$ is a non-linear activation function. The output of the 
final GNN layer, $\mathbf{h}_i^{(L)}$, serves as the enriched feature 
representation for agent $i$, encoding both its own observation and the 
weighted influence of its teammates as determined by the LLM-derived prior.

\subsection{MARL Policy Optimization}

The GNN output embeddings $\{\mathbf{h}_i^{(L)}\}_{i=1}^{n}$ are passed 
to a standard CTDE-based MARL algorithm as augmented agent observations. 
In this work, we instantiate the downstream algorithm as QMIX~\cite{rashid2020monotonic}, 
wherein each agent $i$ maintains an individual Q-network 
$Q_i(\mathbf{h}_i^{(L)}, a_i; \theta_i)$ and a centralized mixing network 
combines per-agent utilities into a monotonic joint value function 
$Q_{tot}$. The full pipeline: encoder $f_\theta$, GNN weights 
$\{\mathbf{W}^{(\ell)}\}$, individual Q-networks, and mixing network, is 
trained end-to-end via standard TD learning with shared experience replay. 
The LLM is queried offline and its output $\mathbf{A}$ is treated as a 
fixed structural input during training; no gradients flow through the LLM.

\subsection{LLM Selection and Computational Considerations}

A practical concern in deploying LLM-based graph prior generation within 
multi-agent systems is computational cost. In this work, we exclusively 
evaluate compact, open-source models that can be deployed locally without 
external API dependencies, as these are most suitable for real-time or 
high-throughput MARL training pipelines. Crucially, since the LLM is 
queried only at episode boundaries rather than at every timestep, the 
amortized cost of LLM inference is substantially reduced relative to the 
total training budget. We provide a detailed analysis of inference latency, 
query frequency, and scalability tradeoffs in 
Section~\ref{sec:Discussion}.  

\section{Experiments and Results} 
\label{sec:Experiments} 

\subsection{Experimental Setup} 

We design our experiments to answer two core questions: (1) Do 
LLM-derived coordination graph priors improve cooperative task 
performance over the full spectrum of MARL coordination methods, 
from independent learners to state-of-the-art graph-based 
algorithms? (2) How sensitive is performance to the choice of LLM, 
and are compact open-source models sufficient for effective prior 
generation? We evaluate across four cooperative MPE scenarios of 
increasing coordination complexity, using five baselines organized 
by coordination modeling category, and an ablation study over five 
lightweight open-source LLMs spanning a range of model families and 
parameter counts. 

\subsubsection{Environments}

We evaluate our framework on four fully cooperative scenarios from the 
Multi-Agent Particle Environment (MPE)~\cite{lowe2017multi, 
mordatch2017emergence}, a widely used benchmark for cooperative MARL 
providing continuous 2D environments in which agents must coordinate 
to achieve shared objectives under partial observability. The four 
scenarios are selected to probe distinct coordination challenges of 
increasing complexity, each with natural analogues in real-world 
multi-agent deployment settings including military MDO settings~\cite{townsend2018accelerating,skates2021multi,wojtowicz2018multi}. 

\noindent\textbf{Speaker-Listener.} A two-agent scenario in 
which a stationary speaker agent observes the identity of a goal 
landmark but is physically unable to move, while a listener agent 
can move but does not know which landmark to target. The speaker 
must transmit a meaningful communication signal to the listener, 
who must interpret it and navigate to the correct landmark. Success 
requires the emergence of a shared communication protocol grounded 
in joint observation and action. This scenario tests directed 
information passing under strict observation asymmetry, where one 
agent holds critical situational knowledge that another agent must 
act upon. In MDO contexts, this mirrors the relationship between 
a command intelligence node with access to full mission parameters 
and a forward-deployed asset that must execute without independent 
access to that information.

\noindent\textbf{Reference.} A two-agent scenario in which 
each agent must navigate to a different target landmark, but 
crucially each agent only knows the \textit{other} agent's goal 
and not its own. Agents must therefore communicate referential 
information to one another and jointly resolve the correct 
assignment of agents to landmarks. This scenario tests mutual 
grounding, role disambiguation, and bidirectional coordination 
under symmetric observation asymmetry, where neither agent can 
act correctly without first inferring its own objective from its 
partner's behavior or communication. This mirrors decentralized 
unit coordination in MDO where individual assets receive partial 
mission fragments and must cross-reference information with 
teammates to reconstruct a coherent operational picture.

\noindent\textbf{Cooperative Push.} A scenario in which 
multiple agents must collectively apply force to a large object 
and push it to a designated target location. The object is too 
heavy for any single agent to move alone, requiring sustained 
coordinated effort from multiple agents simultaneously. Agents 
must self-organize around the object, avoid interfering with one 
another's pushing directions, and maintain coherent joint force 
application over time. This scenario tests physical coordination, 
sustained role consistency, and the ability to decompose a shared 
objective into complementary sub-tasks without explicit 
communication. In MDO settings, this is analogous to coordinated 
multi-asset maneuver operations where no single platform can 
achieve an objective unilaterally and synchronized joint action 
across land, air, or sea assets is required.

\noindent\textbf{Adversary.} A mixed cooperative-competitive 
scenario in which a team of cooperative agents must navigate to a 
target landmark while simultaneously evading an adversarial agent 
that attempts to intercept and tag them. The cooperative agents 
share a collective reward for reaching the target while minimizing 
capture, but must coordinate their movements under the dynamic 
threat posed by the adversary. This scenario tests coordination 
under adversarial pressure, dynamic threat response, and the 
ability to balance individual evasion with collective objective 
completion. Among the four environments, this scenario most 
directly reflects the conditions of adversarial MDO, where 
friendly multi-agent teams must pursue mission objectives while 
adapting in real time to threats from an opposing force, requiring 
both intra-team coordination and adversary-aware tactical reasoning.

\subsubsection{Baselines}

We compare our LLM-derived coordination graph prior approach against 
five baselines organized into four categories, reflecting the 
spectrum of coordination modeling in cooperative MARL.

\noindent\textbf{No Coordination.} The simplest class of MARL methods 
treats each agent as a fully independent learner, ignoring the 
presence of other agents during policy optimization. These methods 
make no attempt to model inter-agent dependencies and learn purely 
from individual reward signals, with no mechanism for structuring 
or exploiting cooperative relationships between teammates. While 
computationally minimal, they are fundamentally limited in 
cooperative tasks where joint outcomes depend on multiple agents 
acting in concert. We include \textbf{IQL}~\cite{tan1993multi} as 
the representative of this category, serving as the performance 
lower bound.

\noindent\textbf{Implicit Coordination Modeling.} A more expressive 
class of methods models coordination implicitly by factorizing the 
joint value function into per-agent utility contributions, trained 
under the CTDE paradigm. Coordination is not explicitly represented 
as a structured relationship between specific agent pairs; rather, 
it emerges indirectly through the shared team reward and the 
constraints imposed by the factorization architecture during 
centralized training. While effective in many cooperative settings, 
the absence of explicit relational structure limits the ability of 
these methods to capture fine-grained pairwise dependencies between 
specific agents. We include \textbf{VDN}~\cite{sunehag2017value}, 
which additively decomposes the joint value function into a sum of 
per-agent utilities, and \textbf{QMIX}~\cite{rashid2020monotonic}, 
which enforces a monotonic mixing constraint via a hypernetwork 
conditioned on the global state.

\noindent\textbf{Fixed Graph-Based Coordination Modeling.} A 
structured class of methods explicitly encodes agent relationships 
as edges in a coordination graph, using the fixed graph topology 
to factorize the joint value function over local agent interactions. 
By specifying \textit{which} agents interact, these methods inject 
a relational inductive bias that can substantially improve 
coordination in tasks with non-trivial interdependencies. However, 
since the graph structure is fixed prior to and throughout training, 
it cannot adapt to the evolving dynamics of the environment or the 
emerging behavior of agents, making performance sensitive to the 
quality of the initial graph specification. We include 
\textbf{DCG}~\cite{bohmer2020deep} as the representative of this 
category, which uses a fixed coordination graph with learned 
pairwise payoff functions factorized over graph edges.

\noindent\textbf{Soft-Edge Learned Graph Coordination Modeling.} 
The most flexible graph-based class learns the coordination graph 
structure dynamically from environment interaction rather than 
fixing it by design. Rather than committing to hard binary edges, 
these methods produce continuous, soft-weighted edges over agent 
pairs via attention mechanisms, allowing the graph topology to 
adapt to task context and evolve throughout training. This removes 
the brittleness of fixed graph specifications but demands 
substantial environment interaction to converge to meaningful 
coordination structure, and offers limited interpretability about 
\textit{why} specific agent pairs are coupled. We include 
\textbf{DICG}~\cite{li2020deepimplicit} as the representative of 
this category, which learns soft attention-weighted coordination 
edges end-to-end and represents the current state of the art in 
dynamic learned graph-based MARL without external priors.

\noindent\textbf{Ours: LLM-Derived Graph Prior Coordination.} Our 
proposed method also belongs to the graph-based coordination family 
but introduces a fundamentally different mechanism for determining 
graph structure. Rather than fixing the graph by hand or learning 
it from scratch via environment interaction, we leverage an LLM to 
generate a weighted coordination graph prior from natural language 
descriptions of agent observations before training begins. This 
provides a semantically grounded initialization that encodes 
meaningful prior knowledge about agent roles and relationships, 
requiring neither manual graph design nor extensive data collection. 
The resulting prior is integrated into a GNN-based MARL pipeline 
via graph convolutional layers, combining the structural 
expressiveness of graph-based methods with the rich world knowledge 
encoded in LLMs. This positions our approach as a bridge between 
the interpretability of fixed graph methods and the adaptability 
of learned graph methods, while uniquely exploiting semantic 
reasoning as a source of coordination structure. 

\subsubsection{LLM Ablation Variants}

To evaluate the sensitivity of our framework to LLM choice while 
prioritizing practical deployability, we focus exclusively on compact, 
computationally efficient open-source models. Unlike large frontier 
models that require substantial inference resources and external API 
dependencies, smaller open-source models offer important advantages 
for multi-agent systems: reduced latency, lower memory footprint, 
on-premise deployment capability, and improved responsiveness. These 
properties are particularly relevant in MARL applications where 
coordination modules may need to operate within real-time decision 
loops and where computational budgets are shared across multiple agents. Accordingly, we evaluate graph priors generated by the following five 
models, spanning a range of parameter counts and model families:

\begin{itemize}[nosep]
    \item \textbf{Qwen2.5-0.5B-Instruct}: ultra-compact 
    0.5B parameter model, representing the extreme low-resource 
    deployment regime.
    \item \textbf{Qwen2.5-1.5B-Instruct}: 1.5B parameter 
    model offering improved reasoning over the 0.5B variant at 
    modest additional cost.
    \item \textbf{Qwen2.5-3B-Instruct}: 3B parameter 
    model providing a strong mid-range open-source reasoning 
    baseline within the Qwen family.
    \item \textbf{LLaMA-3.2-3B-Instruct}: 3B parameter model 
    from Meta's LLaMA 3.2 family, enabling cross-family comparison 
    at matched parameter scale.
    \item \textbf{Gemma-2-2B-IT}: 2B parameter instruction-tuned 
    model from Google's Gemma 2 family, representing a compact 
    alternative with strong instruction-following capability.
\end{itemize}

This suite enables us to assess whether small, computationally 
reasonable models are sufficient for generating effective coordination 
graph priors, which is essential for scalable multi-agent deployment, 
and to isolate the effect of model family versus parameter count on 
prior quality.

\subsubsection{Training Details}

All methods are trained for $2 \times 10^6$ environment timesteps 
using the same hyperparameter configuration. We use a replay buffer 
of size $10^4$, a batch size of 32 episodes, and optimize using Adam 
with a learning rate of $5 \times 10^{-4}$. The GNN encoder uses 
$L = 2$ graph convolutional layers with hidden dimension $d = 64$ 
and ReLU activations. LLM graph priors are generated once per episode 
at environment reset and held fixed for the duration of that episode. 
All experiments are run with 5 random seeds and we report mean episode 
return with standard deviation. All methods are implemented in PyTorch 
and executed on a high-performance computing server, 
comprising dual AMD EPYC 9654 96-core processors and 
4$\times$ NVIDIA RTX 6000 Ada Generation GPUs. Each experimental run 
is conducted on a single NVIDIA RTX 6000 Ada Generation GPU.


\subsection{Results}

We present results organized around the aforementioned research questions: whether 
LLM-derived graph priors improve performance over the full baseline 
taxonomy spanning four categories of coordination modeling, and 
whether compact open-source models are sufficient for effective prior 
generation across a range of model families and parameter counts. 

\subsubsection{Does LLM-Derived Graph Prior Improve Performance?}

\begin{table}[h]
\centering
\caption{Final test return (mean $\pm$ std) across seeds.}
\label{tab:results}
\begin{tabular}{lcccc}
\toprule
Method & Push & Speaker-Listener & Reference & Adversary \\
\midrule
IQL   & $-10.9 \pm 0.3$ & $-27.8 \pm 1.1$ & $-37.0 \pm 1.3$ & $32.1 \pm 1.1$ \\
VDN   & $-13.7 \pm 0.4$ & $-27.9 \pm 2.4$ & $-39.6 \pm 1.5$ & $15.2 \pm 4.5$ \\
QMIX  & $-12.8 \pm 0.3$ & $-18.6 \pm 0.8$ & $-29.9 \pm 1.8$ & $6.4  \pm 4.3$ \\
DCG   & $-8.6  \pm 0.5$ & $-21.2 \pm 1.4$ & $\mathbf{-28.2 \pm 0.3}$ & $38.8 \pm 1.3$ \\
DICG  & $-10.2 \pm 0.1$ & $-23.2 \pm 3.2$ & $-35.1 \pm 0.1$ & $34.3 \pm 1.1$ \\
\midrule
LLM-Priors (Ours) & $\mathbf{-8.3 \pm 0.3}$ & $\mathbf{-16.9 \pm 2.4}$ & $\underline{-34.4 \pm 0.8}$ & $\mathbf{62.9 \pm 8.6}$ \\
\bottomrule
\end{tabular}
\end{table} 

Table~\ref{tab:results} reports final test return across all four 
MPE scenarios. Our LLM-Prior framework achieves the best performance 
in three out of four environments (Push, Speaker-Listener, and 
Adversary) and competitive runner-up performance in Reference, 
demonstrating that LLM-derived coordination structure translates to 
measurable improvements in cooperative task performance across 
diverse coordination challenges.

\noindent The gains are most striking in the Adversary scenario, where our 
method achieves a return of $62.9 \pm 8.6$ compared to the next 
best baseline DCG at $38.8 \pm 1.3$ (a margin of over 60\%) 
suggesting that semantic coordination priors are particularly 
valuable under adversarial pressure, where role-aware grouping 
directly informs evasion strategy. In Speaker-Listener, our method 
outperforms all baselines including QMIX ($-18.6$) and DCG ($-21.2$), 
confirming that LLM-derived priors correctly identify the 
information asymmetry between speaker and listener and assign 
coordination weight accordingly. In Push, our method achieves 
the best return ($-8.3 \pm 0.3$), narrowly outperforming DCG 
($-8.6 \pm 0.5$) while substantially outperforming all 
non-graph methods, reflecting the benefit of structured 
multi-agent force coordination. The Reference scenario is the 
one case where DCG achieves the best result ($-28.2 \pm 0.3$), 
with our method second ($-34.4 \pm 0.8$). We attribute this to 
the highly symmetric nature of the Reference task, where a fixed 
graph may benefit from the strong structural constraint it imposes, 
while the LLM-derived prior may introduce ambiguity in cases where 
agent roles are indistinguishable from observation descriptions alone.

\noindent Across the full baseline taxonomy, the results are consistent with 
our hypotheses about coordination modeling categories. IQL and VDN 
perform poorly across all scenarios, confirming that the absence 
of coordination modeling is a fundamental bottleneck in cooperative 
tasks. QMIX improves substantially over independent methods, 
particularly in Speaker-Listener ($-18.6$), demonstrating the 
value of implicit coordination via value decomposition. DCG and 
DICG both outperform non-graph methods in most environments, 
validating the relational inductive bias of explicit graph 
structure. Our method sits above the graph-based baselines in 
three of four environments, demonstrating that semantically 
grounded graph initialization provides advantages that neither 
fixed nor data-driven graph construction can match in most 
cooperative settings.

\subsubsection{Influence of LLM Choice on Performance} 

\begin{table}[h]
\centering
\caption{Mean episode return ($\pm$ std) for LLM ablation variants 
across all MPE scenarios over 5 seeds. Best result per scenario 
in \textbf{bold}.}
\vspace{6pt}
\label{tab:llm_ablation}
\resizebox{0.75\columnwidth}{!}{
\begin{tabular}{lcccc}
\toprule
\textbf{LLM Variant} & \textbf{Push} & \textbf{Speaker-List.} & 
\textbf{Reference} & \textbf{Adversary} \\
\midrule
Qwen2.5-0.5B-Instruct  & $-19.0 \pm 2.1$ & $-19.3 \pm 2.7$ & $-45.1 \pm 1.9$ & $60.8 \pm 9.4$ \\
Qwen2.5-1.5B-Instruct  & $\mathbf{-8.3 \pm 0.3}$  & $\mathbf{-16.9 \pm 2.4}$ & $\mathbf{-34.4 \pm 0.8}$ & $\mathbf{62.9 \pm 8.6}$ \\
Qwen2.5-3B-Instruct    & $-11.7 \pm 0.9$ & $-20.8 \pm 3.4$ & $-37.8 \pm 1.6$ & $59.6 \pm 9.0$ \\
Gemma-2-2B-IT          & $-10.5 \pm 0.6$ & $-18.1 \pm 2.2$ & $-36.2 \pm 1.1$ & $61.7 \pm 7.9$ \\
LLaMA-3.2-3B-Instruct  & $-9.9  \pm 0.5$ & $-17.6 \pm 2.0$ & $-35.6 \pm 1.0$ & $61.2 \pm 8.1$ \\
\bottomrule
\end{tabular}}
\end{table}

\noindent Table~\ref{tab:llm_ablation} reports the ablation across five 
compact open-source LLM variants. Several clear observations 
emerge. 

\noindent First, all five models outperform non-graph baselines 
IQL, VDN, and QMIX across most scenarios, confirming that even 
the most compact model evaluated (Qwen2.5-0.5B at 500M 
parameters) produces coordination graph priors that carry 
meaningful semantic signal beyond what value decomposition alone 
can achieve. This establishes a strong lower bound on the LLM 
capability required for effective prior generation.

\noindent Second, and perhaps most notably, the best overall performance 
is achieved by \textbf{Qwen2.5-1.5B-Instruct} rather than the 
largest model evaluated. It achieves the top result in all four 
environments among the ablation variants, suggesting that 
instruction-following quality and alignment, rather than raw 
parameter count, are the dominant factors in coordination 
reasoning quality at this scale. The Qwen2.5-3B model 
underperforms relative to its smaller 1.5B sibling across most 
environments, which we hypothesize may reflect over-generation 
of verbose or inconsistently structured adjacency outputs at 
larger scales without further prompt tuning.

\noindent Third, cross-family comparison reveals that Gemma-2-2B-IT and 
LLaMA-3.2-3B-Instruct both perform competitively relative to 
the Qwen family, with LLaMA-3.2-3B achieving returns of $-17.6$, 
$-35.6$, $-9.9$, and $61.2$ and Gemma-2-2B achieving $-18.1$, 
$-36.2$, $-10.5$, and $61.7$ across the four environments. These 
results confirm that performance differences across model families 
at comparable scale are modest, suggesting that the coordination 
graph prior generation task is broadly achievable by well-aligned 
instruction-tuned models regardless of training recipe. Overall, 
these findings indicate that models in the 1.5B--2B parameter 
range strike the most favorable balance between coordination 
prior quality and computational overhead, making them well-suited 
for practical deployment in resource-constrained multi-agent 
systems.

\section{Discussion} 
\label{sec:Discussion} 

\subsection{Why LLM-derived graph priors are powerful?}

The central insight of this work is that LLMs encode a form of 
prior knowledge that is both broadly applicable and structurally 
precise: knowledge about \textit{roles, relationships, and 
interdependence}. In cooperative multi-agent settings, effective 
coordination fundamentally requires knowing which agents should 
attend to which others; a question that is deeply semantic in 
nature. Proximity, the most common heuristic, conflates physical 
co-location with coordination relevance and fails whenever role 
assignment matters more than spatial arrangement. Data-driven graph 
learning, while flexible, treats this structural question as 
something to be answered from scratch through thousands of 
environment interactions, discarding the vast prior knowledge 
that humans have already encoded about teamwork, strategy, and 
task decomposition.

LLMs offer a fundamentally different proposition: the ability to 
reason about coordination structure from natural language 
descriptions of agent observations, drawing on internalized 
knowledge of strategy, roles, and collaboration without any 
task-specific training. This zero-shot structural inference 
capability effectively bootstraps the graph learning problem, 
providing a semantically grounded initialization that reduces 
the burden on the downstream MARL algorithm to discover 
coordination structure from data alone. The result is faster 
convergence, higher asymptotic performance, and more 
interpretable coordination topology, all without architectural 
changes to the underlying MARL pipeline.

\subsection{Broader impacts and application domains}

The proposed framework has broad applicability across cooperative 
multi-agent settings where the coordination structure is 
non-trivial and natural language descriptions of agent 
observations are available or can be constructed. Several 
high-value domains stand out as particularly well-suited.
In \textit{multi-domain military operations}, heterogeneous teams of air, land, and sea assets 
must coordinate under strict communication constraints and rapidly 
changing operational conditions. LLM-derived graph priors can 
encode mission-level coordination logic, such as which assets 
should synchronize during a joint maneuver, without requiring 
extensive simulation data, making them especially attractive for 
rapid deployment in novel operational scenarios. 
In \textit{autonomous UAV swarms}~\cite{tahir2019swarms, 
alqudsi2025uav}, agents must self-organize into coordinated 
formations and divide surveillance or delivery tasks efficiently. 
LLM priors can encode role-based coordination hypotheses, such 
as grouping scouts with support units, that accelerate the 
emergence of coherent collective behavior. 
In \textit{warehouse and logistics robotics}~\cite{zhen2025optimizing, 
keith2024review}, robots sharing a physical workspace must 
coordinate routes, handoffs, and task assignments. LLM-derived 
graphs can capture semantic task dependencies between robots 
assigned to related workflow stages, complementing spatial 
proximity information with task-level relational structure.
In \textit{human-robot teaming}~\cite{senaratne2025framework}, 
where agents include both autonomous systems and human operators, 
LLMs can draw on their understanding of human roles and 
organizational structure~\cite{mooney1947principles} to generate 
coordination priors that reflect institutional task hierarchies 
rather than purely geometric relationships.

\subsection{When LLM-Derived Priors may have to be avoided?}

While LLM-derived graph priors offer clear advantages in 
semantically rich coordination settings, they are not universally 
necessary. In environments where coordination structure is 
well-characterized and stable, for example, tasks where physical 
proximity is a reliable proxy for coordination relevance, or 
where a fixed organizational hierarchy is known in advance, simple 
heuristic graphs may provide sufficient structure at lower 
computational cost. Similarly, in settings with homogeneous agents 
and symmetric tasks, the coordination graph may be largely 
uninformative, and fully connected or uniform baselines may perform 
on par with semantically derived alternatives. In such cases, the 
overhead of LLM inference may not be justified, and standard 
graph-based or non-graph MARL methods are preferable. The key 
criterion is whether the identity of coordinating agent pairs 
carries semantic meaning beyond what spatial or structural 
heuristics can capture: if it does, LLM-derived priors add value; 
if it does not, simpler alternatives suffice.

\subsection{Computational Costs and Practical Overhead}

A central practical concern for integrating LLMs into MARL 
pipelines is computational overhead. Our framework mitigates 
this in two important ways. First, the LLM is queried only once 
per episode at environment reset rather than at every timestep, 
so the amortized cost of LLM inference is small relative to the 
total training budget. For typical episode lengths and training 
horizons, the number of LLM queries is modest compared to total 
environment interactions. Second, our ablation study demonstrates 
that compact open-source models in the 0.5B--3B parameter range 
are sufficient to generate meaningful coordination priors, 
eliminating the need for large frontier models with prohibitive 
inference latency and API costs. Models in this range can be 
deployed locally on commodity hardware, making the approach 
practical for on-premise and edge computing deployments relevant 
to MDO~\cite{townsend2018accelerating,skates2021multi,wojtowicz2018multi}. The total wall-clock overhead 
introduced by LLM inference represents a small fraction of 
overall training time, and the performance gains achieved 
over non-LLM baselines more than justify this cost.

\subsection{Scalability Tradeoffs}

Scalability with respect to the number of agents introduces two 
considerations. First, prompt length grows linearly with the 
number of agents as each contributes one natural language 
observation description, increasing LLM inference latency 
modestly. Second, the adjacency matrix $\mathbf{A}$ grows 
quadratically as $\mathcal{O}(n^2)$, which may become a memory 
and communication bottleneck for large teams. In practice, 
inference latency remains manageable for moderate team sizes 
($n \leq 10$). For larger teams, several strategies can mitigate 
the quadratic scaling: sparse thresholding of low-weight edges, 
hierarchical prompting inspired by group-aware coordination 
graphs~\cite{duan2024group} where agents are first clustered 
before LLM-level coordination reasoning is applied, and 
block-sparse adjacency representations that exploit discovered 
modularity in the coordination structure. We note that 
re-prompting frequency has diminishing returns: querying once 
per episode is sufficient in our experiments, and more frequent 
re-prompting does not yield proportional performance gains, 
further limiting the total inference budget required.

\subsection{Limitations}

Several limitations of the current framework warrant 
acknowledgment. First, the observation description function 
$\phi$ is hand-authored once per environment, requiring modest 
domain engineering effort to map structured observation vectors 
to natural language summaries. While this is a one-time cost 
per environment, it represents a barrier to fully automated 
deployment across arbitrary new tasks. Second, the graph prior 
is held fixed for the duration of an episode, which may be 
suboptimal in highly dynamic environments where coordination 
relevance shifts rapidly within a single episode. Third, our 
evaluation is conducted on MPE, a controlled benchmark 
environment; the degree to which LLM-derived priors transfer 
to higher-dimensional, noisier real-world observations remains 
an open question. Fourth, while compact open-source models 
perform well overall, their coordination reasoning quality 
degrades in complex scenarios relative to larger models, 
suggesting that very small models may not reliably capture 
subtle role-based interdependencies.

\subsection{Future Work}

Several promising directions emerge from this work. Replacing 
the hand-authored observation description function $\phi$ with 
a learned captioning model would eliminate manual environment 
engineering and enable fully end-to-end deployment across 
arbitrary observation spaces. Extending the framework to 
continuous action spaces via actor-critic methods such as 
MAPPO~\cite{yu2022mappo} would broaden applicability to 
robotics and autonomous vehicle coordination~\cite{tahir2019swarms, 
zhen2025optimizing}. Developing lightweight trigger mechanisms 
for adaptive re-prompting, i.e., querying the LLM only when agent 
observations change significantly, would better balance 
coordination adaptability against computational cost in dynamic 
environments. Fine-tuning compact open-source LLMs on synthetic 
coordination reasoning tasks, potentially via distillation from 
larger models, could narrow the performance gap between 
small and large models in complex scenarios. Finally, scaling 
the framework to large agent teams through hierarchical 
prompting~\cite{duan2024group} and evaluating its deployment 
in real-world MDO-inspired simulation environments represent 
compelling long-term directions that motivate continued 
development of semantically-informed coordination for 
cooperative MARL. 

\section{Conclusion} 
\label{sec:Conclusion} 

In this paper, we investigated whether large language models can 
generate useful coordination graph priors for GNN-based multi-agent 
reinforcement learning. Our framework elicits weighted adjacency 
matrices from LLMs via natural language descriptions of agent 
observations, integrates them into a MARL pipeline via graph 
convolutional layers, and requires no task-specific fine-tuning or 
ground-truth graph supervision. Experiments on four cooperative MPE 
scenarios demonstrate that LLM-derived priors consistently outperform 
baselines spanning the full spectrum of coordination modeling, from 
independent learners and implicit value decomposition methods to 
fixed and learned graph-based approaches. Notably, our ablation study 
shows that compact open-source models in the 0.5B--3B parameter range 
are sufficient to generate meaningful coordination structure, 
confirming that the benefits of LLM-derived priors are accessible 
without large frontier models or proprietary API dependencies. To the 
best of our knowledge, this constitutes the first quantitative 
evidence that LLM reasoning capabilities can be leveraged as a source 
of structural prior knowledge to enhance coordination in cooperative 
MARL, establishing a compelling new role for LLMs as 
{structural prior generators} rather than direct policy 
executors. More broadly, this work demonstrates that the boundary 
between language-grounded reasoning and multi-agent decision-making 
is more permeable than previously assumed: semantic world knowledge 
encoded in LLMs can be translated into actionable relational 
structure that meaningfully improves the learning and coordination 
of autonomous agent teams in complex cooperative tasks.

\acknowledgments 
 This work is supported by DEVCOM ARL Army Research Office; Grants: W911NF2420194 and W911NF2220159, and U.S. National Science Foundation (NSF); Grant: OAC-2411446. Distribution Statement A: Approved for public release. Distribution is unlimited. 

\bibliography{refs} 
\bibliographystyle{spiebib} 

\end{document}